\pdfoutput=1

\documentclass[11pt]{article}
\usepackage{acl}

\usepackage{times}
\usepackage{latexsym}

\usepackage[T1]{fontenc}
\usepackage[utf8]{inputenc}

\usepackage{amssymb}%
\usepackage{pifont}%
\newcommand{\cmark}{\ding{51}}%
\newcommand{\xmark}{\ding{55}}%

\usepackage{microtype}

\newif\ifwithappendix
\withappendixfalse

\newif\ifappendixshown
\appendixshowntrue

\usepackage[T1]{fontenc}
\usepackage[utf8]{inputenc}
\usepackage[textsize=tiny]{todonotes}
\usepackage{graphicx}
\usepackage{booktabs}
\usepackage{latexsym}
\usepackage{microtype}

\usepackage{footnote}
\makesavenoteenv{tabular}
\makesavenoteenv{table}
\makesavenoteenv{table*}

\usepackage{todonotes}

\newcommand\minput[1]{%
  \input{#1}%
  \ifhmode\ifnum\lastnodetype=11 \unskip\fi\fi}

\usepackage{hyperref}
\usepackage{xstring}

\newcommand{\noqa}[1]{}
\newcommand{\noqall}[1]{}

\usepackage{amsmath}
\usepackage{booktabs}

\usepackage{multirow}

\usepackage[normalem]{ulem}
\usepackage{placeins}
\usepackage{makecell}

\title{Adapting LLMs for Minimal-edit Grammatical Error Correction}

\author{ Ryszard Staruch \\
  Adam Mickiewicz University \\
  Center for Artificial Intelligence \\
  \texttt{ryszard.staruch@amu.edu.pl} \\ \And
  Filip Graliński \\
  Adam Mickiewicz University \\
  \texttt{filipg@amu.edu.pl} \\
  \AND \bf Daniel Dzienisiewicz \\
  Adam Mickiewicz University \\
  \texttt{dzienis@amu.edu.pl} \\}

\begin{document}
\maketitle
\begin{abstract}

Decoder-only large language models have shown superior performance in the fluency-edit English Grammatical Error Correction, but their adaptation for minimal-edit English GEC is still underexplored. To improve their effectiveness in the minimal-edit approach, we explore the error rate adaptation topic and propose a novel training schedule method. Our experiments set a new state-of-the-art result for a single-model system on the BEA-test set. We also detokenize the most common English GEC datasets to match the natural way of writing text. During the process, we find that there are errors in them. Our experiments analyze whether training on detokenized datasets impacts the results and measure the impact of the usage of the datasets with corrected erroneous examples. To facilitate reproducibility, we have released the source code used to train our models.\footnote{\href{https://github.com/richardxoldman/llms-for-minimal-gec}{github.com/richardxoldman/llms-for-minimal-gec}}

\end{abstract}

\section{Introduction}

Grammatical Error Correction (GEC) is a Natural Language Processing task that covers the detection and correction of errors in texts. Current state-of-the-art models are either Sequence-to-Edit (Seq2Edit) models (encoder-only Transformers) that are trained to tag erroneous tokens and apply proper changes to them \cite{omelianchuk-etal-2020-gector}, or Sequence-to-Sequence (Seq2Seq) models (encoder-decoder Transformers) that are trained to generate the correct version of a given text \cite{rothe-etal-2021-simple}.

Over the years, two main directions have been established in GEC research: minimal-edit GEC and fluency-edit GEC \cite{bryant-etal-2023-grammatical}. The former focuses on applying only the minimal changes necessary to make the text grammatical and error-free. In contrast, fluency-edit GEC goes beyond minimal corrections to achieve native-language fluency.

Current decoder-only large language models (LLMs) achieve state-of-the-art performance on many NLP tasks. Instruction-tuned LLMs are able to produce high-quality texts and correct errors in the zero-shot approach, even without task-specific fine-tuning \cite{davis-etal-2024-prompting}. On the JFLEG dataset \cite{napoles-etal-2017-jfleg}, which is a fluency-edit GEC dataset, the GPT3 and GPT4 models are capable of producing state-of-the-art results \cite{loem-etal-2023-exploring, coyne2023analyzingperformancegpt35gpt4}. LLMs were also used by the winners of the recent multilingual grammatical error correction shared task -- MultiGEC-2025 \cite{masciolini-etal-2025-multigec}.

However, for a minimal-edit GEC, there is only one research work that reports better results compared to other solutions on English minimal-edit GEC benchmarks \cite{liang-etal-2025-edit}. The problem encountered by LLMs can be explained by the phenomenon of overcorrection \cite{fang2023chatgpthighlyfluentgrammatical}.

To further explore LLMs adaptation for minimal-edit GEC, there is a need to find solutions that could allow LLMs to produce more strict outputs. \citet{junczys-dowmunt-etal-2018-approaching} by exploring the error rate adaptation topic show that neural network based solutions need more erroneous examples. Their experiments show that removing the correct examples leads to greater recall. Our intuition is that for modern LLMs, which are able to produce fluent corrections with high linguistic freedom even in the zero-shot manner, the opposite direction is needed, as there is a need for higher precision.

\citet{sun-wang-2022-adjusting} propose a method for a precision-recall
trade-off that requires beam-search decoding, which increases
computational resources and inference time compared to greedy
decoding. To overcome this issue, we propose a novel training schedule method to control the precision-recall trade-off during training instead of inference. Our method allows for the application of standard greedy decoding during inference without the need for external tools or algorithms to control the inference process.

Since LLMs are trained on raw texts and existing GEC datasets are available in word-tokenized (henceforth referred to as ''tokenized'') format \cite{bryant-etal-2023-grammatical}, it forces models to switch from working on raw texts to tokenized texts. 

Another case that would require detokenized texts is any work that leverages probability distributions for language models, for example the Scribendi Score reference-less metric \cite{islam-magnani-2021-end}. 

To solve this issue, we detokenize the most common GEC datasets and verify whether training models on detokenized texts leads to better results. The detokenization process involved the usage of the LLM, during which we discovered that even the most popular datasets contain errors in annotations. We make the detokenized datasets available to the public to make them accessible to other researchers\footnote{\href{https://github.com/richardxoldman/detokenized-gec-datasets}{github.com/richardxoldman/detokenized-gec-datasets}}.

In summary, our contributions in this work are as follows:

\begin{itemize}
\item The LLM that achieves the state-of-the-art single-model system on the BEA-19 Shared Task test set.
\item The study of error rate adaptation in the context of LLMs.
\item The novel training schedule method that enables control of the precision-recall trade-off during training.
\item The detokenization of the most common English GEC datasets, and the detailed analysis of annotation errors in them.
\end{itemize}

\section{Datasets and their detokenization}
\label{sec:datasets}

\begin{table}[]
  \centering
   ...to a cafe and \sout{and} I drank a drink.   \\
   I recommend you \textbf{to} practise any sport...   \\
   She is \textbf{one} of the ones that... \\
   Sometimes we go to part\sout{y}ies in the city. \\
   ...and I was very \textbf{happy} to hug him because I miss him...  \\
  \caption{Examples of changes in target texts made during
    detokenization process by the Llama 3 70b model. Deletions are highlighted with a strikethrough, and insertions are highlighted in bold.}
  \label{tab:expllama}
\end{table}

The most common GEC datasets for English are available in a tokenized format due to evaluation tools that use the M2 format \cite{dahlmeier-ng-2012-better} such as ERRANT \cite{bryant-etal-2017-automatic}. LLMs are trained on raw texts, so the tokenization process forces them to switch to the tokenized text and also to learn the tokenization process. To solve this issue, we detokenize FCE-train \cite{yannakoudakis-etal-2011-new}, W\&I+LOCNESS train and dev part (hereafter, we refer to the train split of this dataset as BEA-train, the dev split as BEA-dev and the test split as BEA-test) \cite{bryant-etal-2019-bea} CoNLL-2014-test \cite{ng-etal-2014-conll}, and JFLEG datasets — these are the datasets we decided to use in our work, as they are one of the most commonly used GEC resources \cite{bryant-etal-2023-grammatical}. The statistics about them are given in the Appendix.

For the FCE-train, BEA-train, and BEA-dev datasets, the source texts were available in the raw format (the only work needed was to properly split them line by line). To detokenize the target texts of these datasets, we used the Sacremoses Detokenizer\footnote{\href{https://pypi.org/project/sacremoses/}{pypi.org/project/sacremoses/}}, but it did not correctly detokenize all the examples. 

To improve the detokenization process, we leveraged the Llama-3.1-70b-Instruct model (denoted as Llama 3 70b), where the model task was only to detokenize the target text. We included a source text that is properly detokenized in the prompt to help the model in the detokenization process. The prompt is given in the Appendix.

In order to detokenize the CoNLL-2014 input texts, we had to properly split paragraphs at the sentence level, which are available in SGML format. We did this using a simple Python script with split rules and then manually adjusted examples that were not properly handled by the script.

\begin{table*}
  \centering
  \begin{tabular}{cc|cccc|c}
    \hline
    \textbf{Dataset} & \textbf{modified} & \textbf{essential} &
    \textbf{optional} & \textbf{erroneous} & \textbf{not assessable} &
    \makecell{\textbf{wrong} \\\textbf{annotations}\\(estimated\\ lower bound)} \\
    \hline
    \textbf{BEA-dev} &  6.52\% & 80.77\% & 2.80\% & 12.59\% & 3.85\% & 5.22\% \\
    \textbf{BEA-train} &  6.22\% & 78.67\% & 4.90\% & 9.80\% & 6.64\% & 4.89\% \\
    \textbf{FCE-train} &  8.42\% & 71.68\% & 12.24\% & 12.24\% & 3.85\%
    & 6.04\% \\
    \hline
  \end{tabular}
  \caption{Details for annotations to examples changed by the Llama 3 70b model.}
  \label{tab:annotations}
\end{table*}

For the JFLEG dataset we only had to detokenize inputs of the dataset, since the dataset has only dev and test splits. Due to the small size of the JFLEG dataset, we used Sacremoses Detokenizer and then manually adjusted the texts.

It should be emphasized that our work does not affect the examples in the test sets. The source texts for both the BEA-test and the CoNLL-2014-test were unchanged. The BEA-test target texts are hidden on the CodaLab platform and are not available publicly. There was no need to detokenize the CoNLL-2014-test target texts, since the scoring script uses the M2 format to compute the results. It makes outcomes based on our version of the datasets fully comparable to the previous research. 

The results reported on our version of the BEA-dev dataset may differ slightly from those reported by other researchers due to the changes described in Section \ref{sec:incorrect-annotations}, but are intended to select the most promising model, not to report the final results.

\subsection{Incorrect annotations in datasets} \label{sec:incorrect-annotations}

In less than 10\% of the examples, the Llama 3 70b model, when used for detokenization, occasionally modified the text beyond simply removing spaces in the correct version of the text. Table~\ref{tab:expllama} shows examples of differences between the target texts in the dataset and the changes made by the Llama 3 70b model. Our initial investigation showed that those changes are mostly errors that were not corrected by a human annotator. Given this, we decided to do a manual annotation of such samples.

For our annotation purposes, the considered sentences were assigned four labels: {\em essential}, {\em optional}, {\em erroneous} and {\em not assessable}. 
  
The {\em essential} label was assigned to sentences in
which corrections were necessary and actually contributed to
improving their accuracy. 

The {\em optional} label was attributed to
sentences in which the corrections made were not necessary, as their original versions were considered correct as well (e.g. sentences originally written in capital letters, which were then changed to lower case).

The {\em erroneous} label refers to situations where the corrections either do not fix the original mistakes in the sentences or create new mistakes in sentences that were already correct.

Finally, the {\em not assessable} label is used to mark corrections for which the quality, for various reasons, cannot be assessed by the annotator.

For BEA-dev, all examples (284) modified by the Llama 3 70b model were
verified, whereas for the other two datasets, random samples of the
same size (284 examples) were checked. The results of the annotation process are shown in Table \ref{tab:annotations}.

\begin{table*}
    \centering
    \begin{tabular}{lcccccc}
    \toprule
        \multirow{2}{*}{\textbf{Model}} & \multirow{2}{*}{\textbf{Size}} & \multirow{2}{*}{\textbf{Setup}} & \multicolumn{3}{c}{\textbf{BEA-dev}} & \multicolumn{1}{c}{\textbf{JFLEG-dev}} \\
    \cmidrule(lr){4-7}
& & & P & R & F\textsubscript{0.5} & GLEU \\
    \midrule
        Qwen 2.5 & 1.5B & detokenized-filtered & 57.90 & 42.10 & 53.86 & 56.10\\
        Qwen 2.5 & 1.5B & tokenized-filtered & 59.00 & 38.48 & 53.31 & 56.17\\
        Qwen 2.5 & 1.5B & detokenized-full & 57.86 & 42.75 & 54.04 & 56.22  \\
        Qwen 2.5 & 1.5B & tokenized-full & 59.92 & 37.79 & 53.63 & 56.01 \\ \hline
        Llama 3 Small & 3B & detokenized-filtered & 63.34 & 47.52 & 59.39 & 57.42\\
        Llama 3 Small & 3B & tokenized-filtered & 63.31 & 47.29 & 59.29 & 57.58\\
        Llama 3 Small & 3B & detokenized-full & 63.04 & 48.32 & 59.42 & 57.56 \\
        Llama 3 Small & 3B & tokenized-full & 62.61 & 46.22 & 58.46 & 56.96 \\
        \hline
        Gemma 2  & 9B & detokenized-filtered  & 68.84 & 56.40 & 65.93 & 58.70 \\
        Gemma 2  & 9B & tokenized-filtered & 68.84 & 55.90 & 65.79 & 58.99 \\
        Gemma 2  & 9B & detokenized-full & 69.07 & 57.13 & 66.30 & 58.72\\
        Gemma 2 & 9B & tokenized-full & 69.86 & 55.67 & 66.47 & 58.40\\

    \bottomrule
    \end{tabular}
    \caption{Results for different dataset processing setups.}
    \label{tab:exp_diff}
\end{table*}

\subsection{Detokenization impact}

To verify whether the detokenization process and the modification of examples by the Llama 3 70b model have an impact on the GEC models, we decided to train the LLMs on the FCE-train and the BEA-train datasets in four different processing setups:
\begin{enumerate}
\item \textbf{detokenized-filtered}: Detokenized datasets \textbf{excluding} examples modified by the Llama 3 70b model.
\item \textbf{tokenized-filtered}: Tokenized datasets corresponding to the examples that remained unmodified in the detokenized version.
\item \textbf{detokenized-full}: Detokenized datasets \textbf{including} all examples, both modified and unmodified.
\item \textbf{tokenized-full}: Tokenized datasets corresponding to the full set of detokenized examples (original, untouched datasets).

\end{enumerate}

Please note that \textbf{tokenized-*} setups refer to the original examples ''as is'', without any modifications introduced by the Llama 3 70b model.

The \textbf{detokenized-filtered} setup compared to the \textbf{tokenized-filtered} setup shows whether the detokenization process has an impact on the models' performance, since both models are fine-tuned on the same examples with the same hyperparameter setup. The details about the hyperparameters are given in the Appendix.

The \textbf{*-full} setups against the \textbf{*-filtered} setups show whether the changes made by the Llama 3 70b model in the datasets have an impact on the results, because the \textbf{detokenized-full} setup contains the modified examples by the Llama 3 70b model, whereas the \textbf{tokenized-full} setup contains all the original examples (also the erroneous ones). Again, the number of training examples is the same, but the difference lies in the quality of the annotations in examples that were changed by the Llama 3 70b model.

All models were trained for one epoch on the FCE-train dataset and then for one epoch on the BEA-train dataset. In this and subsequent experiments, we report the results for the BEA-dev and JFLEG-dev datasets, since these datasets give a view for both minimal-edit and fluency-edit GEC. Table \ref{tab:exp_diff} presents the results for 3 different LLMs of different sizes: Qwen2.5-1.5B-Instruct (denoted as Qwen 2.5), Llama-3.2-3B-Instruct (denoted as Llama 3 Small) and gemma-2-9b-it (denoted as Gemma 2). 

\begin{table}
  \centering
  \begin{tabular}{cccc}
    \hline
    \textbf{Dataset} & \textbf{M} & \textbf{R} & \textbf{U} \\
    \hline
    \textbf{BEA-dev} &  50.74\% & 38.87\% & 10.39\% \\
    \textbf{BEA-train} &  46.93\% & 40.28\% & 12.79\% \\
    \textbf{FCE-train} &  61.33\% & 31.96\% & 6.71\% \\
    \hline
  \end{tabular}
  \caption{Details about the operations performed by the Llama 3 70b model. The labels stand for: Missing, Replacement and Unnecessary.}
  \label{tab:operations}
\end{table}

\subsection{Results analysis}

The results show that LLMs can learn the tokenized version of the texts and in some cases even achieve better metric scores compared to the models trained on the detokenized texts. We can see that there are no clear gains in terms of F\textsubscript{0.5} score from using the detokenized version of datasets. 

The transition from the \textbf{tokenized-filtered} to the \textbf{tokenized-full} setup increases precision in each experiment but lowers recall and GLEU values. In all cases, transition from the \textbf{detokenized-filtered} setup to the \textbf{detokenized-full} setup improves recall and slightly improves the GLEU score. It shows that the changes made by the Llama 3 70b model result in outputs with higher linguistic freedom, which is expected, since the most common change made by the Llama 3 70b model is the Missing operation (Table~\ref{tab:operations}), while using the original sentences makes the models produce more strict outputs.

We can also see that the size of the models significantly impacts the results. Therefore, for the next experiments we will further explore the Gemma 2 model, as it is the best performing model. Although Gemma 2 achieves the best F\textsubscript{0.5} score on the \textbf{tokenized-full} setup, the next experiments will be performed on the detokenized version of the datasets, as they contain corrected erroneous annotations. The other reason is that our systems can be used in the work of other researchers who need a model that produces detokenized output. It would be also simply practical in terms of using the system in the environment where the output does not require removing the unnecessary spaces.

\section{Overcorrection problem}

\begin{figure*}
\includegraphics[width=\linewidth]{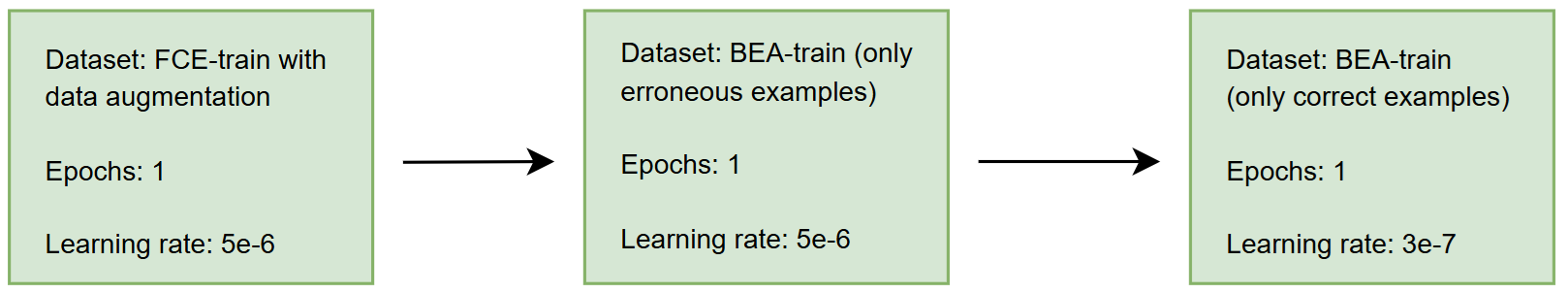}
\caption{Visualization of the fine-tuning process for our best performing Gemma 2 model on the BEA-dev dataset.}
\label{fig:screenshot}
\end{figure*}

In the minimal-edit GEC task, the goal is to find and correct only those parts of the texts that are clearly erroneous, without making further improvements to their fluency. Due to the pre-training goal of LLMs, which is to maximize the probability of the next token, and the flexibility they gain from instruction fine-tuning process, LLMs tend to produce more fluent output. While this characteristic may be advantageous for fluency-edit GEC, the objective of minimal-edit GEC is to apply only the minimal necessary corrections.

Standard minimal-edit GEC benchmarks, which are based on texts written by English language learners, put a greater weight on precision than on recall, because suggesting an incorrect change is considered more negative than ignoring an error \cite{ng-etal-2014-conll}. Therefore, a proper adaptation of the model is needed to correct errors with high precision.

For the Chinese minimal-edit GEC, \citet{yang-quan-2024-alirector} proposes an alignment model which is used to filter only minimal corrections from the initial correction, which may be fluent.

\begin{table*}
  \centering
  \begin{tabular}{ccccccc}
    \toprule
        \multirow{2}{*}{\textbf{Dataset processing approach}} & \multicolumn{2}{c}{\textbf{Erroneous sentences}} & \multicolumn{3}{c}{\textbf{BEA-dev}} & \multicolumn{1}{c}{\textbf{JFLEG-dev}} \\
    \cmidrule(lr){2-7}
& FCE-train & BEA-train & P & R & F\textsubscript{0.5} & GLEU \\
\midrule
    ONLY-ERRONEOUS & 100\% & 100\% & 60.74 & 58.79 & 60.34 & 58.16 \\
    UNCHANGED & 65.43\% & 69.02\% & 68.99 & 57.12 & 66.24 & 58.73 \\
    AUGMENTED & 39.55\% & 40.83\% & 71.42 & 53.42 & \textbf{66.92} & 58.21 \\ \hline
  \end{tabular}
  \caption{Results for the dev sets for the experiment with our data augmentation method.}
  \label{tab:dataset_process}
\end{table*}

One of the most recent works proposes the novel method for LLM fine-tuning, Edit-Wise Preference Optimization (EPO) that fits the minimal-edit GEC task better than the standard supervised fine-tuning (SFT) approach \cite{liang-etal-2025-edit}. In our work, we explore the SFT approach with a focus on the datasets rather than the different training approaches, and show that proper data preprocessing or training schedule can lead to the successful minimal-edit LLM model.

\section{Data augmentation}

During GEC model fine-tuning, datasets play a crucial role in the whole process. One of the most important attributes of the GEC datasets is the error rate. The common practice for neural models that are trained from scratch is to remove unedited pairs \cite{chollampatt10.5555/3504035.3504741, kiyono-etal-2019-empirical}, because for these models there is a need for improved recall. 

Large language models produce fluent output with high recall, which may suggest that removing unedited pairs for LLMs is unnecessary and could worsen the results. Furthermore, it may be possible that providing additional unedited pairs could improve minimal-edit error correction for LLMs.

To provide more real examples that may not be fluent, but are still acceptable, we propose a data augmentation method to split each example (consisting of source text and corrected text) into two pairs. The new pair is created by using the corrected text as both the source and target text. For example, the sentence pair ``Alice have a cat.'' and ``Alice has a cat.'' can be split into the following examples: 
\begin{itemize}
    \item $\text{Alice have a cat.}\rightarrow\text{Alice has a cat.}$
    \item $\text{Alice has a cat.}\rightarrow\text{Alice has a cat.}$
\end{itemize}

Our method can be applied to any dataset and does not require any additional models/tools to extend a given dataset.

\section{Training schedule}

Current approaches to GEC training scheduling consist of dividing data into 2 or 3 groups based on data quality and then training a model in the correct order, from the lowest quality data to the highest \cite{bout-etal-2023-efficient}. We follow this approach, but to control the precision-recall trade-off, we propose to extend it even further. 

In the final stage (with the highest quality dataset -- in our case it is BEA-train dataset), we split the data into two groups. The first group contains only erroneous texts, whereas the second group contains only correct examples. During the stage, we first train the model on the first group (only erroneous examples), and then we train the model on the second group (only correct examples) with lower learning rate. Figure~\ref{fig:screenshot} shows the step-by-step training schedule for the best performing model on the BEA-dev set.

Our intuition behind this approach is that a model first learns how to correct errors and later is tuned to understand which examples are correct but in some cases not perfectly fluent. During the last stage, when the model is fine-tuned only on correct examples, the model only learns to not apply corrections to texts. 

Choosing a proper learning rate value (or number of examples) enables controlling the precision-recall trade-off in LLMs, as lowering learning rate should make the model learn not to correct more smoothly while still being able to correct the errors in texts.  

\begin{table*}
  \centering
  \begin{tabular}{cccccc}
    \toprule
        \multirow{2}{*}{\textbf{Learning rate}} & \multirow{2}{*}{\makecell{\textbf{FCE-train} \\ \textbf{Augmented}}} & \multicolumn{3}{c}{\textbf{BEA-dev}} & \multicolumn{1}{c}{\textbf{JFLEG-dev}} \\
    \cmidrule(lr){3-6}
& & P & R & F\textsubscript{0.5} & GLEU \\
\midrule
    1e-7  & \xmark  & 65.90 & 58.18 & 64.19 & 58.58  \\
    1e-7  & \cmark  & 65.10 & 58.33 & 63.62 & 58.60 \\ \hline
    2e-7  & \xmark  & 69.30 & 56.05 & 66.17 & 58.64  \\
    2e-7  & \cmark  & 69.22 & 56.40 & 66.21 & 58.66 \\ \hline
    2.5e-7  & \xmark  & 70.94 & 53.73 & 66.67 & 58.47  \\
    2.5e-7  & \cmark  & 70.96 & 54.40 & 66.89 & 58.28   \\ \hline
    3e-7  & \xmark  & 73.63 & 48.72 & 66.80 & 57.60  \\
    3e-7  & \cmark  & 73.52 & 50.10 & \textbf{67.23} & 57.90   \\ \hline
    3.5e-7  & \xmark  & 75.81 & 44.92 & 66.65 & 56.74  \\
    3.5e-7  & \cmark  & 75.38 & 46.82 & 67.18 & 57.35   \\ \hline
    4e-7  & \xmark  & 77.49 & 40.15 & 65.34 & 55.48  \\
    4e-7  & \cmark  & 76.74 & 43.49 & 66.57 & 56.15   \\ \hline
    
    5e-7  & \xmark  & 79.74 & 24.79 & 55.29 & 50.26  \\
    5e-7  & \cmark  & 78.88 & 31.78 & 60.85 & 52.91   \\ \hline
  \end{tabular}
  \caption{Results for the dev sets for the experiment with our training schedule method.}
  \label{tab:exp2}
\end{table*}

\section{Experiments}

\subsection{Data augmentation experiments} \label{subsec:data-aug-exp}

To test whether the addition of unedited pairs can positively affect LLMs in the minimal-edit GEC task, we train the Gemma 2 model\footnote{For the data augmentation and training schedule experiments we also tested the gemma-2-9b-it-SimPO model and achieved slightly better results, but we decided to use the original Gemma 2 model as our goal is not to maximize the benchmark scores.} with the same hyperparameter setup as in the experiment from Section \ref{sec:datasets} in three different dataset processing approaches:
\begin{itemize}
    \item only erroneous examples (denoted as \textbf{ONLY-ERRONEOUS})
    \item erroneous examples + unedited examples (denoted as \textbf{UNCHANGED})
    \item erroneous examples + unedited examples + unedited examples created from erroneous examples by applying our data augmentation method (denoted as \textbf{AUGMENTED})
\end{itemize}

As in the previous experiment, we first train one epoch on the FCE-train dataset and then one epoch on the BEA-train dataset.

Table \ref{tab:dataset_process} shows the results on the BEA-dev and JFLEG-dev datasets. We can see that unedited examples are needed to improve the LLMs performance. Even on the fluency-edit dataset, the scores are better when unedited pairs are added to the dataset (the \textbf{UNCHANGED} approach). For the \textbf{AUGMENTED} approach, the F\textsubscript{0.5} score is the highest among all approaches, but the GLEU score is lower compared to the \textbf{UNCHANGED} approach. 

This study shows that lowering the error rate in the GEC datasets is a way to make LLMs produce minimal-edit outputs. It also shows that when new solutions are available, such as modern LLMs, approaches or practices from previous research, such as removing unedited pairs, should be reevaluated and tested again.

\subsection{Training schedule experiment}
\label{subsec:train-schedule-exp}

We also carried out an experiment with different learning rate values for the last group (only correct examples) for our training schedule method for the Gemma 2 model. We also test whether applying our data augmentation method for the FCE-train dataset improves the results. 

Note that in this experiment data augmentation method is \textbf{not} applied to the BEA-train dataset.

Table~\ref{tab:exp2} shows how the precision-recall trade-off depends on the learning rate value. It can be observed that even small changes in the learning rate value noticeably influence the trade-off, making the hyperparameter very sensitive.

When applying the data augmentation method for the FCE-train dataset, the BEA-dev set F\textsubscript{0.5} score can be improved compared to the best value achieved in the previous experiment (the \textbf{AUGMENTED} dataset processing approach). 

Although the data augmentation method was designed to enhance precision, we observe that results with data augmentation on the FCE-train have higher recall. In this experiment, we hypothesize that training on the FCE-train provides general GEC knowledge, while fine-tuning on the BEA-train determines the model's behavior in terms of the precision-recall trade-off as model is first fine-tuned on erroneous examples and then on the correct ones.

Figure~\ref{fig:screenshot} shows the complete training process for the model with the highest F\textsubscript{0.5} score.

\begin{table*}
    \centering
    \begin{tabular}{lccccccc}
    \toprule
        \multirow{2}{*}{\textbf{Model}} & \multirow{2}{*}{\textbf{Size}} & \multicolumn{3}{c}{\textbf{CoNLL-2014-test}} & \multicolumn{3}{c}{\textbf{BEA-test}} \\
    \cmidrule(lr){3-8}
        & & P & R & F\textsubscript{0.5} & P & R & F\textsubscript{0.5} \\
    \midrule
        T5 Large \cite{rothe-etal-2021-simple} & 700M  & - & - & 66.04 & - & - & 72.06 \\
        T5 XL \cite{rothe-etal-2021-simple} & 3B  & - & - & 67.65 & - & - & 73.92 \\
        T5 XXL \cite{rothe-etal-2021-simple} & 11B  & - & - & 68.75 & - & - & 75.88 \\
        GECToR \cite{tarnavskyi-etal-2022-ensembling} & 355M  & 74.40 & 41.05 & 64.00 & 80.70 & 53.39 & 73.21 \\
        TemplateGEC \cite{li-etal-2023-templategec} & 770M  & 74.80 & 50.00 & 68.10 & 76.80 & 64.80 & 74.10 \\
        FLAN-T5 XXL \cite{cao-etal-2023-unsupervised} & 11B  & 75.00 & 53.80 & 69.60 & 78.80 & 68.50 & 76.50 \\
        DeCoGLM \cite{li-wang-2024-detection} & 335M  & 75.10 & 49.40 & 68.00 & 77.40 & 64.60 & 74.40 \\
        BART Base \cite{wang-etal-2024-improving-grammatical} & 400M  & 76.20 & 52.20 & 69.80 & 77.70 & 67.50 & 75.40 \\
        Llama-2-13b \cite{omelianchuk-etal-2024-pillars} & 13B  & 77.30 & 45.60 & 67.90 & 74.60 & 67.80 & 73.10 \\ 
        Mistral-7b-EPO \cite{liang-etal-2025-edit} & 7B & 76.71 & 52.56 & \textbf{70.26} & 78.16 & 68.07 & 75.91\\
        \hline
        Gemma 2 Augmentation & 9B  & 73.80 & \textbf{56.16} & 69.43 & 74.86 & \textbf{71.35} & 74.13 \\
        Gemma 2 Training-Schedule & 9B  & 75.74 & 51.47 & 69.24 & 79.87 & 68.90 & 77.41 \\
        Llama-2-13b Training-Schedule & 13B & 71.07 & 50.11 & 65.59 & 74.10 & 67.54 & 72.69 \\
        Gemma 2 (27b) Training-Schedule & 27B  & \textbf{77.38} & 47.88 & 68.89 & \textbf{82.28} & 67.03 & \textbf{78.70} \\
    \bottomrule
    \end{tabular}
    \caption{Single model results for the minimal-edit GEC test sets.}
    \label{tab:minimal}
\end{table*}

\begin{table}[]
\centering
\begin{tabular}{cc}
\hline
\textbf{Model} & \textbf{GLEU} \\ \hline
Source (Uncorrected) & 40.54 \\
Reference (Average) & 62.37 \\ \hline
Conv Seq2Seq \cite{ge2018reachinghumanlevelperformanceautomatic} & 62.42 \\
\makecell{Transformer \\ \cite{stahlberg-kumar-2021-synthetic}} & 64.70 \\
GPT-3.5 \cite{coyne2023analyzingperformancegpt35gpt4} & 63.40 \\
GPT-4 \cite{coyne2023analyzingperformancegpt35gpt4} & \textbf{65.02} \\ \hline
Gemma 2 Augmentation & 63.72 \\
Gemma 2 Training-Schedule & 62.91 \\
Llama-2-13b Training-Schedule & 62.53 \\
Gemma 2 (27b) Training-Schedule & 62.42 \\ \hline
\end{tabular}
\caption{Results for the fluency-edit GEC dataset (JFLEG-test).}\label{tab:fluency}
\end{table}

\subsection{Results on the test datasets}

From each experiment, we choose the most promising model based on its performance on the BEA-dev dataset to evaluate it on the BEA-test, CoNLL-2014-test, and JFLEG-test datasets. In Table~\ref{tab:minimal}, Gemma 2 Augmentation refers to the best model from Section~\ref{subsec:data-aug-exp} (only applying the data augmentation method) and Gemma 2 Training-Schedule refers to the best model from Section~\ref{subsec:train-schedule-exp}. 

Table~\ref{tab:minimal} shows that our model from the training-schedule experiment achieves a new state-of-the-art single model result on the BEA-test dataset and has competitive results with other solutions on the CoNLL-2014-test dataset. It should be noted that our models were trained only on two relatively small datasets, whereas other solutions were trained on a much larger number of examples, except for the Mistal-7b-EPO model.

To get more insights about the impact of the different model selection on the results, we also performed a single experiment with the gemma-2-27b-it and llama-2-13b-chat (Gemma 2 (27b) Training-Schedule and LLama-2-13b Training-Schedule in the tables) models with the same training schedule and hyperparameters as the best performing model on the BEA-dev dataset, so the model training is exactly the same as for the Gemma 2 Training-Schedule model. 

The Llama-2-13b achieves even worse results than these reported by \cite{omelianchuk-etal-2024-pillars}. It can be explained by using different datasets during fine-tuning process. The precision and recall are both worse than those of the Gemma 2 model. This suggests that model size is not the only important factor; other details about the LLM, such as its novelty, architecture, and the dataset used for training, also matter.

The Gemma 2 (27b) achieves even a better score than the best Gemma 2 9b model on the BEA-test set, but it may be slightly overtuned for precision due to the same learning rate value in the final stage with the bigger model, which can be observed in the worse results for the CoNLL-2014-test dataset.

Table~\ref{tab:fluency} shows the results for the JFLEG-test dataset. We can see that even if our models are fine-tuned for minimal-edit GEC, they achieve a higher score than the average of the scores computed for the JFLEG-test references. It suggests that LLMs can find a proper balance between minimal-edit GEC and fluency-edit GEC.

\section{Conclusions}

Our work demonstrates that there are several ways to fine-tune an LLM for minimal-edit grammatical error correction, without the need for pre-training them on a large number of examples. We propose easy-to-implement methods for controlling the precision-recall trade-off during fine-tuning. 

Moreover, we show that choosing a more recent LLM is also an important factor that impacts the overall performance of the model. The Gemma 2 9b model, even as a smaller model achieved much better performance compared to the Llama-2-13b model.

The detokenization process did not improve model performance, but our findings on the errors in the most common GEC datasets show the need for a proper curation of datasets. Our work also shows that LLMs can be effectively used as a detokenization tool.

\FloatBarrier

\section{Limitations}

Our work covers only experiments on English GEC datasets, so it would be beneficial to extend the research to check how LLMs would perform in other languages. We did not conduct experiments on other types of models. It is hard to tell whether our methods would improve the Seq2Seq or Seq2Edit approaches.

The other issue is that we applied only greedy decoding during inference. The results could be even better if different decoding methods were applied. It would also be worth comparing these methods applied in LLMs with the Seq2Seq or Seq2Edit models.

The reusability of the training schedule method is limited by the requirement for extensive learning rate tuning for any different model or dataset due to high sensitivity to minor changes in learning rate.

Obtaining the highest F\textsubscript{0.5} might be considered overfitting for a specific test set and evaluation metric, but in practical terms, the style of grammar correction depends on specific needs, guidelines, etc., so this might be a desired behavior.

Lastly, running our models requires a lot of memory and computational power, so for many people it would be impossible to run them on their devices. Our models may not be practical for everyday use, but they can be used to create synthetic datasets that can be used to train smaller models.

\bibliography{ms}

\begin{thebibliography}{30}
\providecommand{\natexlab}[1]{#1}

\bibitem[{Bout et~al.(2023)Bout, Podolskiy, Nikolenko, and
  Piontkovskaya}]{bout-etal-2023-efficient}
Andrey Bout, Alexander Podolskiy, Sergey Nikolenko, and Irina Piontkovskaya.
  2023.
\newblock \href {https://doi.org/10.18653/v1/2023.emnlp-main.355} {Efficient
  grammatical error correction via multi-task training and optimized training
  schedule}.
\newblock In \emph{Proceedings of the 2023 Conference on Empirical Methods in
  Natural Language Processing}, pages 5800--5816, Singapore. Association for
  Computational Linguistics.

\bibitem[{Bryant et~al.(2019)Bryant, Felice, Andersen, and
  Briscoe}]{bryant-etal-2019-bea}
Christopher Bryant, Mariano Felice, {\O}istein~E. Andersen, and Ted Briscoe.
  2019.
\newblock \href {https://doi.org/10.18653/v1/W19-4406} {The {BEA}-2019 shared
  task on grammatical error correction}.
\newblock In \emph{Proceedings of the Fourteenth Workshop on Innovative Use of
  NLP for Building Educational Applications}, pages 52--75, Florence, Italy.
  Association for Computational Linguistics.

\bibitem[{Bryant et~al.(2017)Bryant, Felice, and
  Briscoe}]{bryant-etal-2017-automatic}
Christopher Bryant, Mariano Felice, and Ted Briscoe. 2017.
\newblock \href {https://doi.org/10.18653/v1/P17-1074} {Automatic annotation
  and evaluation of error types for grammatical error correction}.
\newblock In \emph{Proceedings of the 55th Annual Meeting of the Association
  for Computational Linguistics (Volume 1: Long Papers)}, pages 793--805,
  Vancouver, Canada. Association for Computational Linguistics.

\bibitem[{Bryant et~al.(2023)Bryant, Yuan, Qorib, Cao, Ng, and
  Briscoe}]{bryant-etal-2023-grammatical}
Christopher Bryant, Zheng Yuan, Muhammad~Reza Qorib, Hannan Cao, Hwee~Tou Ng,
  and Ted Briscoe. 2023.
\newblock \href {https://doi.org/10.1162/coli_a_00478} {Grammatical error
  correction: A survey of the state of the art}.
\newblock \emph{Computational Linguistics}, pages 643--701.

\bibitem[{Cao et~al.(2023)Cao, Yuan, Zhang, and
  Ng}]{cao-etal-2023-unsupervised}
Hannan Cao, Liping Yuan, Yuchen Zhang, and Hwee~Tou Ng. 2023.
\newblock \href {https://doi.org/10.18653/v1/2023.emnlp-main.185} {Unsupervised
  grammatical error correction rivaling supervised methods}.
\newblock In \emph{Proceedings of the 2023 Conference on Empirical Methods in
  Natural Language Processing}, pages 3072--3088, Singapore. Association for
  Computational Linguistics.

\bibitem[{Chollampatt and Ng(2018)}]{chollampatt10.5555/3504035.3504741}
Shamil Chollampatt and Hwee~Tou Ng. 2018.
\newblock A multilayer convolutional encoder-decoder neural network for
  grammatical error correction.
\newblock In \emph{Proceedings of the Thirty-Second AAAI Conference on
  Artificial Intelligence and Thirtieth Innovative Applications of Artificial
  Intelligence Conference and Eighth AAAI Symposium on Educational Advances in
  Artificial Intelligence}, AAAI'18/IAAI'18/EAAI'18. AAAI Press.

\bibitem[{Coyne et~al.(2023)Coyne, Sakaguchi, Galvan-Sosa, Zock, and
  Inui}]{coyne2023analyzingperformancegpt35gpt4}
Steven Coyne, Keisuke Sakaguchi, Diana Galvan-Sosa, Michael Zock, and Kentaro
  Inui. 2023.
\newblock \href {https://arxiv.org/abs/2303.14342} {Analyzing the performance
  of gpt-3.5 and gpt-4 in grammatical error correction}.
\newblock \emph{Preprint}, arXiv:2303.14342.

\bibitem[{Dahlmeier and Ng(2012)}]{dahlmeier-ng-2012-better}
Daniel Dahlmeier and Hwee~Tou Ng. 2012.
\newblock \href {https://aclanthology.org/N12-1067} {Better evaluation for
  grammatical error correction}.
\newblock In \emph{Proceedings of the 2012 Conference of the North {A}merican
  Chapter of the Association for Computational Linguistics: Human Language
  Technologies}, pages 568--572, Montr{\'e}al, Canada. Association for
  Computational Linguistics.

\bibitem[{Davis et~al.(2024)Davis, Caines, Andersen, Taslimipoor,
  Yannakoudakis, Yuan, Bryant, Rei, and Buttery}]{davis-etal-2024-prompting}
Christopher Davis, Andrew Caines, O~Andersen, Shiva Taslimipoor, Helen
  Yannakoudakis, Zheng Yuan, Christopher Bryant, Marek Rei, and Paula Buttery.
  2024.
\newblock \href {https://doi.org/10.18653/v1/2024.findings-acl.711} {Prompting
  open-source and commercial language models for grammatical error correction
  of {E}nglish learner text}.
\newblock In \emph{Findings of the Association for Computational Linguistics
  ACL 2024}, pages 11952--11967, Bangkok, Thailand and virtual meeting.
  Association for Computational Linguistics.

\bibitem[{Fang et~al.(2023)Fang, Yang, Lan, Wong, Hu, Chao, and
  Zhang}]{fang2023chatgpthighlyfluentgrammatical}
Tao Fang, Shu Yang, Kaixin Lan, Derek~F. Wong, Jinpeng Hu, Lidia~S. Chao, and
  Yue Zhang. 2023.
\newblock \href {https://arxiv.org/abs/2304.01746} {Is chatgpt a highly fluent
  grammatical error correction system? a comprehensive evaluation}.
\newblock \emph{Preprint}, arXiv:2304.01746.

\bibitem[{Ge et~al.(2018)Ge, Wei, and
  Zhou}]{ge2018reachinghumanlevelperformanceautomatic}
Tao Ge, Furu Wei, and Ming Zhou. 2018.
\newblock \href {https://arxiv.org/abs/1807.01270} {Reaching human-level
  performance in automatic grammatical error correction: An empirical study}.
\newblock \emph{Preprint}, arXiv:1807.01270.

\bibitem[{Islam and Magnani(2021)}]{islam-magnani-2021-end}
Md~Asadul Islam and Enrico Magnani. 2021.
\newblock \href {https://doi.org/10.18653/v1/2021.emnlp-main.239} {Is this the
  end of the gold standard? a straightforward reference-less grammatical error
  correction metric}.
\newblock In \emph{Proceedings of the 2021 Conference on Empirical Methods in
  Natural Language Processing}, pages 3009--3015, Online and Punta Cana,
  Dominican Republic. Association for Computational Linguistics.

\bibitem[{Junczys-Dowmunt et~al.(2018)Junczys-Dowmunt, Grundkiewicz, Guha, and
  Heafield}]{junczys-dowmunt-etal-2018-approaching}
Marcin Junczys-Dowmunt, Roman Grundkiewicz, Shubha Guha, and Kenneth Heafield.
  2018.
\newblock \href {https://doi.org/10.18653/v1/N18-1055} {Approaching neural
  grammatical error correction as a low-resource machine translation task}.
\newblock In \emph{Proceedings of the 2018 Conference of the North {A}merican
  Chapter of the Association for Computational Linguistics: Human Language
  Technologies, Volume 1 (Long Papers)}, pages 595--606, New Orleans,
  Louisiana. Association for Computational Linguistics.

\bibitem[{Kiyono et~al.(2019)Kiyono, Suzuki, Mita, Mizumoto, and
  Inui}]{kiyono-etal-2019-empirical}
Shun Kiyono, Jun Suzuki, Masato Mita, Tomoya Mizumoto, and Kentaro Inui. 2019.
\newblock \href {https://doi.org/10.18653/v1/D19-1119} {An empirical study of
  incorporating pseudo data into grammatical error correction}.
\newblock In \emph{Proceedings of the 2019 Conference on Empirical Methods in
  Natural Language Processing and the 9th International Joint Conference on
  Natural Language Processing (EMNLP-IJCNLP)}, pages 1236--1242, Hong Kong,
  China. Association for Computational Linguistics.

\bibitem[{Li and Wang(2024)}]{li-wang-2024-detection}
Wei Li and Houfeng Wang. 2024.
\newblock \href {https://doi.org/10.18653/v1/2024.acl-long.96}
  {Detection-correction structure via general language model for grammatical
  error correction}.
\newblock In \emph{Proceedings of the 62nd Annual Meeting of the Association
  for Computational Linguistics (Volume 1: Long Papers)}, pages 1748--1763,
  Bangkok, Thailand. Association for Computational Linguistics.

\bibitem[{Li et~al.(2023)Li, Liu, Wang, Gong, Wong, Gao, Huang, and
  Zhang}]{li-etal-2023-templategec}
Yinghao Li, Xuebo Liu, Shuo Wang, Peiyuan Gong, Derek~F. Wong, Yang Gao, Heyan
  Huang, and Min Zhang. 2023.
\newblock \href {https://doi.org/10.18653/v1/2023.acl-long.380}
  {{T}emplate{GEC}: Improving grammatical error correction with detection
  template}.
\newblock In \emph{Proceedings of the 61st Annual Meeting of the Association
  for Computational Linguistics (Volume 1: Long Papers)}, pages 6878--6892,
  Toronto, Canada. Association for Computational Linguistics.

\bibitem[{Liang et~al.(2025)Liang, Yang, Gao, and Quan}]{liang-etal-2025-edit}
Jiehao Liang, Haihui Yang, Shiping Gao, and Xiaojun Quan. 2025.
\newblock \href {https://aclanthology.org/2025.coling-main.229/} {Edit-wise
  preference optimization for grammatical error correction}.
\newblock In \emph{Proceedings of the 31st International Conference on
  Computational Linguistics}, pages 3401--3414, Abu Dhabi, UAE. Association for
  Computational Linguistics.

\bibitem[{Loem et~al.(2023)Loem, Kaneko, Takase, and
  Okazaki}]{loem-etal-2023-exploring}
Mengsay Loem, Masahiro Kaneko, Sho Takase, and Naoaki Okazaki. 2023.
\newblock \href {https://doi.org/10.18653/v1/2023.bea-1.18} {Exploring
  effectiveness of {GPT}-3 in grammatical error correction: A study on
  performance and controllability in prompt-based methods}.
\newblock In \emph{Proceedings of the 18th Workshop on Innovative Use of NLP
  for Building Educational Applications (BEA 2023)}, pages 205--219, Toronto,
  Canada. Association for Computational Linguistics.

\bibitem[{Masciolini et~al.(2025)Masciolini, Caines, Clercq, Kruijsbergen,
  Kurfal{\i}, S{\'a}nchez, Volodina, and
  {\"O}stling}]{masciolini-etal-2025-multigec}
Arianna Masciolini, Andrew Caines, Orph{\'e}e~De Clercq, Joni Kruijsbergen,
  Murathan Kurfal{\i}, Ricardo~Mu{\~n}oz S{\'a}nchez, Elena Volodina, and
  Robert {\"O}stling. 2025.
\newblock \href {https://aclanthology.org/2025.nlp4call-1.1/} {The
  {M}ulti{GEC}-2025 shared task on multilingual grammatical error correction at
  {NLP}4{CALL}}.
\newblock In \emph{Proceedings of the 14th Workshop on Natural Language
  Processing for Computer Assisted Language Learning}, pages 1--33, Tallinn,
  Estonia. University of Tartu Library.

\bibitem[{Napoles et~al.(2017)Napoles, Sakaguchi, and
  Tetreault}]{napoles-etal-2017-jfleg}
Courtney Napoles, Keisuke Sakaguchi, and Joel Tetreault. 2017.
\newblock \href {https://aclanthology.org/E17-2037/} {{JFLEG}: A fluency corpus
  and benchmark for grammatical error correction}.
\newblock In \emph{Proceedings of the 15th Conference of the {E}uropean Chapter
  of the Association for Computational Linguistics: Volume 2, Short Papers},
  pages 229--234, Valencia, Spain. Association for Computational Linguistics.

\bibitem[{Ng et~al.(2014)Ng, Wu, Briscoe, Hadiwinoto, Susanto, and
  Bryant}]{ng-etal-2014-conll}
Hwee~Tou Ng, Siew~Mei Wu, Ted Briscoe, Christian Hadiwinoto, Raymond~Hendy
  Susanto, and Christopher Bryant. 2014.
\newblock \href {https://doi.org/10.3115/v1/W14-1701} {The {C}o{NLL}-2014
  shared task on grammatical error correction}.
\newblock In \emph{Proceedings of the Eighteenth Conference on Computational
  Natural Language Learning: Shared Task}, pages 1--14, Baltimore, Maryland.
  Association for Computational Linguistics.

\bibitem[{Omelianchuk et~al.(2020)Omelianchuk, Atrasevych, Chernodub, and
  Skurzhanskyi}]{omelianchuk-etal-2020-gector}
Kostiantyn Omelianchuk, Vitaliy Atrasevych, Artem Chernodub, and Oleksandr
  Skurzhanskyi. 2020.
\newblock \href {https://doi.org/10.18653/v1/2020.bea-1.16} {{GECT}o{R} {--}
  grammatical error correction: Tag, not rewrite}.
\newblock In \emph{Proceedings of the Fifteenth Workshop on Innovative Use of
  NLP for Building Educational Applications}, pages 163--170, Seattle, WA, USA
  → Online. Association for Computational Linguistics.

\bibitem[{Omelianchuk et~al.(2024)Omelianchuk, Liubonko, Skurzhanskyi,
  Chernodub, Korniienko, and Samokhin}]{omelianchuk-etal-2024-pillars}
Kostiantyn Omelianchuk, Andrii Liubonko, Oleksandr Skurzhanskyi, Artem
  Chernodub, Oleksandr Korniienko, and Igor Samokhin. 2024.
\newblock \href {https://aclanthology.org/2024.bea-1.3} {Pillars of grammatical
  error correction: Comprehensive inspection of contemporary approaches in the
  era of large language models}.
\newblock In \emph{Proceedings of the 19th Workshop on Innovative Use of NLP
  for Building Educational Applications (BEA 2024)}, pages 17--33, Mexico City,
  Mexico. Association for Computational Linguistics.

\bibitem[{Rothe et~al.(2021)Rothe, Mallinson, Malmi, Krause, and
  Severyn}]{rothe-etal-2021-simple}
Sascha Rothe, Jonathan Mallinson, Eric Malmi, Sebastian Krause, and Aliaksei
  Severyn. 2021.
\newblock \href {https://doi.org/10.18653/v1/2021.acl-short.89} {A simple
  recipe for multilingual grammatical error correction}.
\newblock In \emph{Proceedings of the 59th Annual Meeting of the Association
  for Computational Linguistics and the 11th International Joint Conference on
  Natural Language Processing (Volume 2: Short Papers)}, pages 702--707,
  Online. Association for Computational Linguistics.

\bibitem[{Stahlberg and Kumar(2021)}]{stahlberg-kumar-2021-synthetic}
Felix Stahlberg and Shankar Kumar. 2021.
\newblock \href {https://aclanthology.org/2021.bea-1.4/} {Synthetic data
  generation for grammatical error correction with tagged corruption models}.
\newblock In \emph{Proceedings of the 16th Workshop on Innovative Use of NLP
  for Building Educational Applications}, pages 37--47, Online. Association for
  Computational Linguistics.

\bibitem[{Sun and Wang(2022)}]{sun-wang-2022-adjusting}
Xin Sun and Houfeng Wang. 2022.
\newblock \href {https://doi.org/10.18653/v1/2022.acl-short.77} {Adjusting the
  precision-recall trade-off with align-and-predict decoding for grammatical
  error correction}.
\newblock In \emph{Proceedings of the 60th Annual Meeting of the Association
  for Computational Linguistics (Volume 2: Short Papers)}, pages 686--693,
  Dublin, Ireland. Association for Computational Linguistics.

\bibitem[{Tarnavskyi et~al.(2022)Tarnavskyi, Chernodub, and
  Omelianchuk}]{tarnavskyi-etal-2022-ensembling}
Maksym Tarnavskyi, Artem Chernodub, and Kostiantyn Omelianchuk. 2022.
\newblock \href {https://doi.org/10.18653/v1/2022.acl-long.266} {Ensembling and
  knowledge distilling of large sequence taggers for grammatical error
  correction}.
\newblock In \emph{Proceedings of the 60th Annual Meeting of the Association
  for Computational Linguistics (Volume 1: Long Papers)}, pages 3842--3852,
  Dublin, Ireland. Association for Computational Linguistics.

\bibitem[{Wang et~al.(2024)Wang, Wang, Liu, Zhu, Wu, and
  Che}]{wang-etal-2024-improving-grammatical}
Yixuan Wang, Baoxin Wang, Yijun Liu, Qingfu Zhu, Dayong Wu, and Wanxiang Che.
  2024.
\newblock \href {https://doi.org/10.18653/v1/2024.findings-acl.647} {Improving
  grammatical error correction via contextual data augmentation}.
\newblock In \emph{Findings of the Association for Computational Linguistics
  ACL 2024}, pages 10898--10910, Bangkok, Thailand and virtual meeting.
  Association for Computational Linguistics.

\bibitem[{Yang and Quan(2024)}]{yang-quan-2024-alirector}
Haihui Yang and Xiaojun Quan. 2024.
\newblock \href {https://doi.org/10.18653/v1/2024.findings-acl.148} {Alirector:
  Alignment-enhanced {C}hinese grammatical error corrector}.
\newblock In \emph{Findings of the Association for Computational Linguistics:
  ACL 2024}, pages 2531--2546, Bangkok, Thailand. Association for Computational
  Linguistics.

\bibitem[{Yannakoudakis et~al.(2011)Yannakoudakis, Briscoe, and
  Medlock}]{yannakoudakis-etal-2011-new}
Helen Yannakoudakis, Ted Briscoe, and Ben Medlock. 2011.
\newblock \href {https://aclanthology.org/P11-1019} {A new dataset and method
  for automatically grading {ESOL} texts}.
\newblock In \emph{Proceedings of the 49th Annual Meeting of the Association
  for Computational Linguistics: Human Language Technologies}, pages 180--189,
  Portland, Oregon, USA. Association for Computational Linguistics.

\end{thebibliography}

\appendix
\newpage\hbox{}\thispagestyle{empty}

\section{Training details}

We trained our <=13b models on a 2xA100 (80GB) GPU setup and the 27b model on a 4xA100 (80GB) GPU setup. We used 4xA100 (80GB) GPU setup to run the Llama 3 70b model for the detokenization process. A single model training took 2-3 hours. The hyperparameter values are described in Table~\ref{tab:hyperparameters}. The following prompt was used during training our models and during inference:

Correct the following text, making only minimal changes where necessary.

\#\#\# Text to correct:

<source text>

\#\#\# Corrected text:

<target text>

\section{Detokenization prompt}
The following prompt was used to detokenize the datasets:

You will receive two texts: source text and corrected text. Corrected text may not have proper spaces. Your task is to remove/add proper spaces to the corrected text. Do not write any comments, just write corrected text with proper spaces.

Source text: <source text>

Corrected text: <target text>

Only change spaces, you must not change punctuation.

\FloatBarrier
\begin{table}
  \centering
  \begin{tabular}{c|c|c}
    \hline
    \textbf{Dataset} & \textbf{\#Examples} & \makecell{\textbf{Erroneous} \\ \textbf{sentences}} \\ 
    \hline
    \textbf{FCE-Train} &  28.4k & 65.43\% \\
    \textbf{BEA-train} &  34.3k & 69.02\% \\
    \textbf{BEA-test} &  4.5k & -- \\
    \textbf{BEA-dev} &  4.4k & 67.36\% \\
    \textbf{CoNLL-2014-test} &  1.3k & 71.90\% \\
    \textbf{JFLEG-dev} & 754 & 95.36\% \\
    \textbf{JFLEG-test} & 747 & 95.31\% \\
    \hline
  \end{tabular}
    \caption{Details of the datasets used in our work. Note that there ratio of erroneous sentences could be different compared to the statistics about the datasets from different research works due to the changes made by the Llama 3 70b model during the detokenization process.}
  \label{tab:datasets}
\end{table}

\begin{table}
  \centering
  \begin{tabular}{c|c}
    \hline
    \textbf{Hyperparameter name} & \textbf{Value}  \\ \hline
   learning rate & 5e-6  \\
   batch size & 4  \\
   gradient accumulation steps & 4 \\
   warmup steps (for each dataset) & 100 \\
   lr scheduler & linear \\
   epochs (for each dataset) & 1 \\
   optimizer & AdamW8bit \\
   weight decay & 0.01 \\
   \hline
  \end{tabular}
  \caption{Hyperparameter values used to train our models.}
  \label{tab:hyperparameters}
\end{table}

\end{document}